\pdfoutput=1

\documentclass[11pt]{article}

\usepackage[final]{acl}

\usepackage{times}
\usepackage{latexsym}
\usepackage{graphicx}
\usepackage{booktabs}
\usepackage{ragged2e}
\usepackage{array, multirow}
\usepackage{booktabs,tabularx}

\usepackage[T1]{fontenc}

\usepackage[utf8]{inputenc}

\usepackage{microtype}

\usepackage{array}
\newcolumntype{C}[1]{>{\centering\let\newline\\\arraybackslash\hspace{0pt}}m{#1}}

\title{``It's \emph{how} you do things that matters'': Attending to Process to Better Serve Indigenous Communities with Language Technologies}

\author{Ned Cooper\thanks{\ \ Work done while at Google Research.} \\
  Australian National University  \\
  \texttt{\small edward.cooper@anu.edu.au} \\\And
  Courtney Heldreth \\
  Google Research, USA \\
  \texttt{\small cheldreth@google.com}
  \\\And
  Ben Hutchinson \\
  Google Research, Australia \\
  \texttt{\small benhutch@google.com}
}

\begin{document}
\maketitle
\begin{abstract}
Indigenous languages are historically under-served by Natural Language Processing (NLP) technologies, but this is changing for some languages with the recent scaling of large multilingual models and an increased focus by the NLP community on endangered languages. This position paper explores ethical considerations in building NLP technologies for Indigenous languages, based on the premise that such projects should primarily serve Indigenous communities. We report on interviews with 17 researchers working in or with Aboriginal and/or Torres Strait Islander communities on language technology projects in Australia. Drawing on insights from the interviews, we recommend practices for NLP researchers to increase attention to the process of engagements with Indigenous communities, rather than focusing only on decontextualised artefacts.
\end{abstract}

\section{Introduction}

In this position paper, we discuss how to ethically build Natural Language Processing (NLP) technologies for Indigenous languages, which have historically been poorly served by NLP. This is a timely question, as we are in the \textsc{unesco} International Decade of Indigenous Languages (2022--2032), and there has been a recent trend towards more NLP technologies processing Indigenous languages. One thread of recent projects has been motivated by scaling large multilingual models to include Indigenous languages, including Māori, Zulu, Igbo, Southern Quechua, Hawaiian, Querétaro Otomi, Navajo, and more \cite[\emph{e.g.},][]{Pratap2023,ImaniGooghari2023, Kudugunta2023}. Another thread of recent projects is driven by threats of language extinction, for example, the six Workshops on the Use of Computational Methods in the Study of Endangered Languages (`ComputEL') held since 2014, and the ACL 2022 Theme Track: \emph{``Language Diversity: From Low-Resource to Endangered Languages''}. Both threads of research are typically based on assumptions that language technologies should be accessible to everyone in their first language(s), and that the availability of those language technologies will promote language use and preservation \cite{Bird2020}.

We start with the premise that NLP for Indigenous languages should primarily serve Indigenous communities. If this is indeed the goal of the NLP community, then we need NLP to be accountable to and benefit Indigenous communities \cite{Schwartz2022}, and to prioritise communities' values and experiences with respect to NLP projects. Prioritising these values and experiences specifically includes considering the context of Indigenous communities within colonised societies \cite{Schwartz2022,Bird2020} and the expressed opinions of those communities around data governance \cite[\emph{e.g.,}][]{Liu2022,Mager2023}. The overarching question for this paper, then, is: \emph{how can NLP better serve Indigenous communities}?

To consider this question, we first review the developing discourse around decolonisation of language technology, along with principles for Indigenous data governance. We then report on interviews with researchers working in or with Aboriginal and/or Torres Strait Islander communities on language technology projects in Australia, the country in which two of the authors live. Drawing on insights from the interviews, we recommend practices for NLP researchers working with Indigenous languages. Overall, we encourage NLP researchers to increase attention to the process of engagements with Indigenous communities, rather than focusing only on decontextualised artefacts.

\section{Background}
Languages can be marginalised in different ways. The NLP research community describes a language as `\emph{low-resource}' when there is insufficient data in that language to train and evaluate statistical and machine learning models \cite{Liu2022}. The poverty-conscious framing of the term `low-resource' has been criticised by \citet{Bird2022}, however, for being colonial and Eurocentric. We prefer the term \emph{under-served} in this paper (echoing, for example, \citealp{Bender2018,Kaffee2018,Armstrong2022,Forbes2022}), as we recognise that a language may be fully constituted in its own ways, while it may not be serviced by dominant NLP tools or techniques. Guided by scholars of marginalisation processes (\emph{e.g.}, \citealp{Bagga-Gupta2017}), we seek to pivot the discussion from `low-resource' languages to how technology communities are \emph{under-serving} language communities.

Languages spoken by few people may additionally be defined as \emph{endangered}---at risk of disappearing due to a lack of speakers \cite{Bromham2022}. However, having few living speakers does not necessarily mean a language is `low-resource' (\emph{e.g.}, Latin has enough data to support Google Translate).

The majority of Indigenous languages---languages spoken in a particular region by Indigenous peoples---are forecast to disappear by the end of this century \cite{Bromham2022}. In practice, most Indigenous languages are endangered due not to any inherent linguistic inferiority, but rather due to the global economic, ideological, military, and nationalistic practices that are constitutive of \emph{colonialism}. 

\subsection{Decolonisation and Language Technology}

Decolonial approaches to addressing marginalisation in technology are primarily motivated by social justice and self-determination \cite{Smith1999}, not only data efficiency \cite{Bird2022}. These approaches encourage researchers to embrace perspectives from and at the margins in order to surface and critique the persistence of colonial relationships in present-day society \cite{Maldonado-Torres2007,Quijano2007,Escobar2018}. According to the literature on decolonisation, there are three broad strategies to enact decolonial agendas in language technology work.

Firstly, decolonial agendas require that we \emph{consider whose interests are served by NLP}. Language technologies are laden with cultural perspectives and assumptions \cite{Awori2016}, and NLP has a ``habit of … technological colonisation'' along with making assumptions about goals and methods \cite{Bird2020}. Research on languages of Indigenous communities must be conducted on their terms \cite{Dourish2020} and research outputs must be primarily relevant to those communities, not only to research communities \cite{Alvarado_Garcia2021}.

Secondly, decolonial agendas encourage us to \emph{question the universality of values} \cite{Mignolo2011,Grosfoguel2007}, in particular, the primacy of Western values over others. This includes questioning methods and utility functions of NLP projects. Assuming all communities want the same language technologies disempowers local communities \cite{Bird2020}. Instead, we must critique the universalising logic of our methods, along with technologies \cite{Dourish2020,Irani2010}. In addition, imagining alternative futures for NLP and under-served languages means defending other perspectives and worldviews \cite{Escobar2018}.

Thirdly, decolonial agendas direct us to \emph{interrogate power dynamics embedded in NLP projects}. Approaches from the Global North are often disconnected from the life experiences of those in the Global South \cite{Alvarado_Garcia2021}. In addition, power asymmetries exist between users and platforms \cite{Couldry2018}, and between different regions of the world \cite{Kwet2019}.

\subsection{Principles for Indigenous Data Governance}

We believe it is critical to consider Indigenous perspectives on language data management. Examples of such perspectives are reflected in the CARE principles of the Global Indigenous Data Alliance \cite{Carroll2020}, the \citet{Maiam_nayri_Wingara2018} Indigenous Data Sovereignty Principles, and the \citet{Te_Mana_Raraunga2016} principles of Māori data sovereignty. These principles grapple with an ongoing tension for Indigenous communities when engaging with language technologists---between maintaining sovereignty over their language data and engaging with technological developments that could benefit language revitalisation efforts. Although each set of principles is distinct, a thematic analysis by the authors revealed some common areas of concern.

\begin{itemize}
    \item \emph{Respect}: Acknowledge and support the rights of people and communities to hold and express different values, norms and aspirations regarding data and technology. This requires listening, and understanding culture.
    \item \emph{Relationships}: Act cooperatively. Build positive, long-term relationships.
    \item \emph{Shared control}: Support data governance and control. Support the exercise of data guardianship using traditional protocols.
    \item \emph{Benefits}: Understand disparate benefits and ensure equitable distribution of benefits. Provide evidence of individual and collective benefits.
\end{itemize}

\section{Insights from Interviews}

Building on the previous section, our focus here narrows to Australia as a case study. Australian Aboriginal and Torres Strait Islander languages are marginalised in multiple ways. There is a scarcity of language technologies, which reflects a much broader technological under-serving of these communities. Indeed, many communities struggle to get reliable and affordable access to the internet \cite{featherstone2023}. Prior to colonisation, there were more than 250 local languages spoken in Australia, though today just over 120 languages are in use or being revitalised and more than 90\% of those are considered endangered \cite{Australian_Government2020}. However, it is not for a lack of internet, data, or NLP technologies that many local languages are endangered or extinct. We cannot ignore the impacts of colonialism---in many cases, language loss is the byproduct of oppression. Local languages were often the target of colonial oppression as those languages sustained identities and connection to Indigenous lands and cultures \cite{Bird2020}. For example, up until the mid-20th century, the Commonwealth Government of Australia forced Aboriginal and Torres Strait Islander people to learn English and forbade them from speaking their own languages in government and mission schools \cite{Rademaker2018}.

To delve deeper into this context, we formulated two research questions guiding a series of interviews with researchers who work in or with Aboriginal and/or Torres Strait Islander communities on speech and language technology projects. Firstly, how should language technologists work with local communities to develop speech and language technologies? Secondly, what is the role of speech and language technologies in sustaining language use by local communities?

We conducted semi-structured, 60-minute interviews with 17 researchers from academia and community-based organisations between October 2022 and June 2023 (see Appendix). Our approach to recruitment combined purposeful and snowball sampling \cite{Palinkas2015}. First, we contacted researchers known to the authors who had published on language technology development projects conducted in or with Aboriginal and/or Torres Strait Islander communities in Australia. We asked our initial interviewees to recommend others for us to contact in a second round, including members of Aboriginal and Torres Strait Islander communities researching their own languages. We contacted all potential interviewees via email, and those who accepted our request completed a consent form. The vast majority of the interviews were conducted via video conference, though a few were conducted in person. All interviews were transcribed and shared with interviewees following the interview for review.

Each interview followed the same general format, though we tailored interview guides to each interviewee and their published work. After asking background questions about the project(s) relevant to the study and the interviewee, we asked each interviewee a series of descriptive questions about how they approach working with communities in language technology development projects. Finally, we asked a series of more open-ended questions prompting interviewees to reflect on the present and future of NLP for Indigenous languages in Australia---for example, exploring projects and activities to focus on and practices to avoid.

Insights from the interviews were derived using thematic analysis, initially by hand and in a second round using coding software. The descriptions and reflections of our interviewees, detailed in the following two sections, shed light on strategies and challenges to enact decolonial agendas and Indigenous data governance principles at the project level. We include anonymised quotes from our interviewees throughout the sections.

\subsection{\emph{How} to work with Aboriginal and Torres Strait Islander communities}

We first asked interviewees how they decide what to work on and who to work with. All interviewees strongly emphasised that speech and language technology projects \emph{``must start with a community need''}, and that recognising such needs requires long-term relationships. The need for translation, for example, often arises where communities or researchers observe something happening across cultures over time. Many interviewees also argued that projects shouldn't start with technology, or solutions. Instead, interviewees encouraged other technologists to demo existing technology and facilitate experimentation with the tools by communities for their languages.

We also asked researchers how they manage relationships with the people they work with. All interviewees emphasised that researchers must clarify to partner communities the mutual benefits of a project at the outset, with some interviewees explicitly mentioning the negotiation of data access rights. Several interviewees noted that community-based work requires researchers to question universal assumptions about the social or cultural factors relevant to technology, and that personal relationships are key to managing those complexities.

Finally, we asked about finishing projects. Most interviewees noted that, though it is important for projects to have an end date, personal relationships between researchers and communities persist. Several interviewees encouraged translating documentation into an accessible form that communities can continue to access (rather than locking up data in bespoke, single purpose tools). Those same interviewees argued that repositories and archives support the sustainability of project outcomes: \emph{``Apps and websites are disposable … store the data in an archival format that is going to persist.''}

\subsection{\emph{What} to work on with Aboriginal and Torres Strait Islander communities}

Most interviewees stated that the primary motivation of the communities they work with for building speech and technologies is the transmission of culture via language: \emph{``Tap into the intrinsic motivation of transmitting life and knowledge down the generations.''} Several interviewees encouraged a \emph{``design for one, then scale''} approach, where researchers collaborate with one community, then scale a \emph{``digital shell''}---a technological template tailored for one community, yet adaptable enough to be customised by others---streamlining early development stages for each new engagement. Others urged technologists to consider the benefits of the production process to communities, to facilitate capacity building in technology development, not only focusing on project outputs like datasets or publications.

In terms of application domains, several interviewees advocated for improving accessibility to archival materials using front-end tools for metadata tagging and information retrieval, especially for audio. Others emphasised the importance of vehicular languages like Aboriginal English, Australian Kriol, and Torres Strait Creole. Interviewees noted that many communities use vehicular languages to participate in the national economy and access education and health systems. Finally, some interviewees encouraged multi-modal work to support signed Aboriginal languages, alongside text and audio.

\section{Recommendations and Conclusion}

To conclude, we propose a set of practices building on the insights from our interviews, along with the decolonial agendas and Indigenous data governance principles outlined earlier. We recognise that Indigenous communities and their languages vary considerably across the world, and the needs of communities in one region might not necessarily reflect those of other regions. However, many Indigenous communities have common experiences with respect to colonialism and its links to research practices \cite{Smith1999}. Therefore, while our interview study was specific to Australia, we put forward these practices for NLP researchers to test and build upon in other regions.

\begin{figure}
    \centering
    \includegraphics[width=6cm]{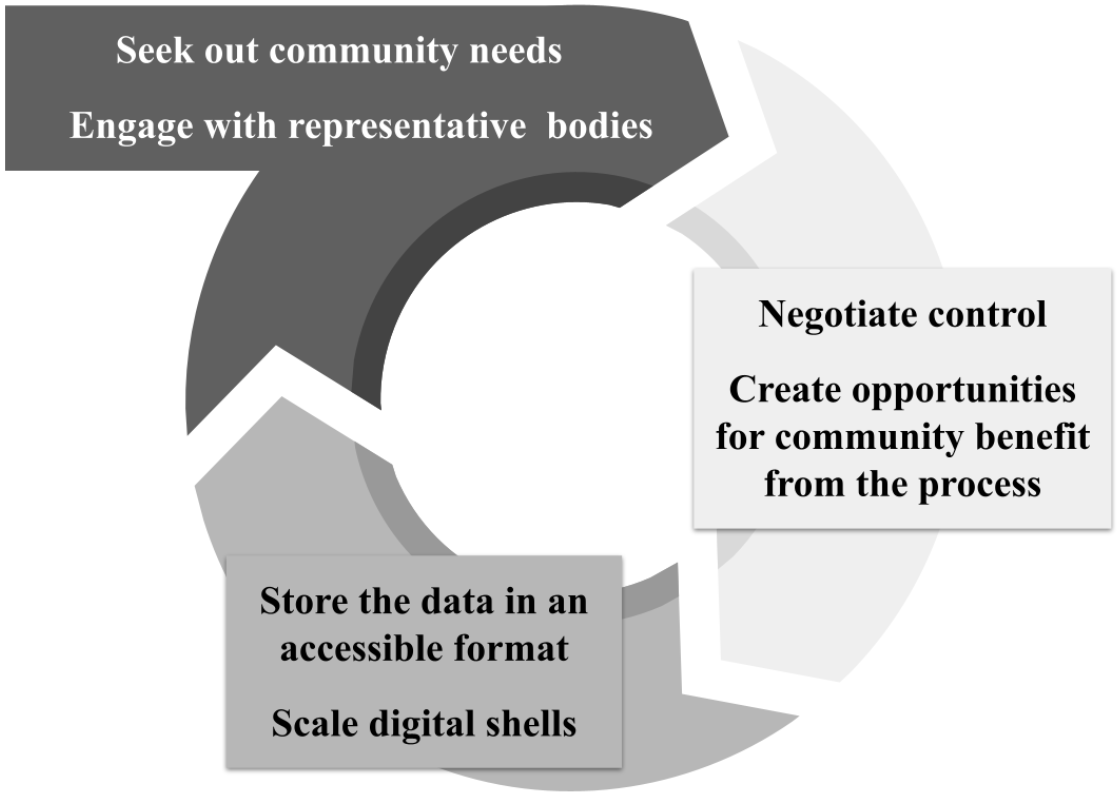}
    \caption{Recommended process for engagement.}
    \label{fig:process}
\end{figure}

The practices grapple with a tension for NLP researchers working with Indigenous languages---between producing work that is relevant to local partner communities and the demands of research communities for projects that scale across many languages. We intend to contribute to the discourse about decolonisation of language technology, not by resolving this tension, but by recommending a cyclical process of engagement to assist researchers to navigate it (Figure~\ref{fig:process}). As \citet{Escobar2018} suggests for design, we argue that the NLP community can engage with marginalisation and dispossession through a greater focus on the process of engagements rather than on artefacts alone.

An ethical process starts by \emph{seeking out community needs}. This means asking communities we wish to partner with about their goals for their languages, and ensuring our efforts are aligned with those goals \cite{Liu2022}. This approach may lead us to focus more on supporting the transmission of cultural knowledge across generations, not only expanding access to products and services. Focusing solely on data collection by communities to develop products and services risks disenfranchising communities. Instead, one approach might be to demo existing technology at community events (\emph{e.g.}, the PULiiMA Indigenous Languages and Technology Conference) and asking how communities can appropriate it for their needs.

\emph{Engaging with community representative bodies} can help researchers establish long-term relationships with community members. While personal relationships between researchers and community members are crucial, engaging through representative bodies offers a distinct advantage in balancing power dynamics. Additionally, these bodies already have established relationships within their communities, allowing researchers to build trust and credibility more rapidly.

Relatedly, we must consider how to \emph{negotiate control over project resources and ongoing relationships}. At the start of community-engaged language technology projects, this involves several steps. Firstly, researchers should schedule time to interrogate power dynamics \cite{Blodgett2020}, which involves recognising the often distinct decision-making processes and communication approaches of researchers and Indigenous community participants, and developing mutually agreed protocols for the project \cite{Cooper2022}. Secondly, it's important to consider how to share power with community partners by recognising Indigenous \mbox{(co-)ownership} of outcomes of data collection efforts (\emph{e.g.}, community ownership of datasets or other intellectual property, and joint publications \cite{janke2021true}).

Where data collection is a component of a project with an Indigenous community, we must consider how \emph{the process of engagement might be an opportunity for community benefit}. In practice, this may involve designing experiences for community members to learn about language technology as part of the process of generating or collecting data, and creating outputs from data collection that are accessible by community members, not only usable by language technologists.

In addition, it is critical to \emph{store and maintain data produced from the project} in a format that community partners can access beyond the project (\emph{e.g.}, archives or repositories). Where researchers also intend to scale projects across languages, we recommend starting small---focusing on one to two communities, then \emph{scaling digital shells} to other contexts (see, for example, \citealp{Richards2019,Foley2018}).

Finally, we also urge the NLP research environment to pay more attention to the process of engaging with Indigenous communities, rather than focusing on de-contextualised model accuracy benchmarks as proxies for utility to communities \cite{Hutchinson2022}. This means including the process of engagement as a core reviewing criterion when processing Indigenous languages, and fostering forums where Indigenous voices can articulate their needs to the NLP community. While NLP research processes can, like NLP artefacts, be a source of harm to communities \cite{Ashurst2022}, such processes can be generative for communities when guided by their methodologies and interests \cite[see \emph{e.g.}, `yarning' as a research method:][]{Rodríguez2021}. Let the process of engagement with Indigenous communities and their voices be the pillars of our research.


\section{Limitations}

Our paper has several limitations. Firstly, our sampling approach may introduce selection bias, as the initial set of researchers we recruited influenced the final group of interviewees. Our approach favoured researchers in our existing networks, which could overlook the perspectives of researchers and community members outside those networks. Secondly, we conducted interviews with researchers working in or with Aboriginal and/or Torres Strait Islander communities; however, the majority of our interviewees were non-Indigenous, and we did not recruit any interviewees from Torres Strait Islander communities. While engaging with researchers aligns with our goal of understanding strategies and challenges for NLP researchers to enact decolonial agendas and Indigenous data governance principles, we have limited insight into the perspectives of Aboriginal and Torres Strait Islander communities. Thirdly, our positions of privilege as researchers affect our interpretation and presentation of themes from the interviews. While we strive for understanding and sensitivity, our perspectives could overlook lived experiences that we might not recognise. This emphasises the importance of considered engagement with Indigenous voices to ensure appropriate representation in NLP research about Indigenous languages.

\section{Acknowledgements}

We would like to thank our interviewees, whose insights and experiences informed this paper, and extend that thanks to the communities collaborating with our interviewees. Additionally, our thanks go to the organisers and participants of the 2023 PULiiMA Indigenous Languages and Technology Conference, for enriching discussions that improved the paper, and to the anonymous reviewers for their suggestions and feedback.

\bibliography{anthology,custom}
\bibliographystyle{acl_natbib}

\appendix
\newpage

\section{Summary of interviewees}
\label{sec:appendix}

\begin{table}[ht]
\centering
\small
\begin{tabular}{p{4cm}>{\RaggedLeft}m{1cm}}
\toprule
Indigenous status & Count \\\midrule
Non-Indigenous & 12 \\
Aboriginal & 5 \\
\bottomrule
\end{tabular}
\caption{Indigenous status of interviewees.}
\label{tab:Indigenous_status}
\end{table}

\begin{table}[ht]
\centering
\small
\begin{tabular}{p{4cm}r}
\toprule
Field of Expertise & Count \\\midrule
Linguistics & 7 \\
Computing & 7 \\
Community-based research & 3\\
\bottomrule
\end{tabular}
\caption{Primary field of expertise of interviewees.}
\label{tab:field_of_expertise}
\end{table}

\begin{table}[ht]
\centering
\small
\begin{tabular}{p{4cm}r}
\toprule
Australian State or Territory & Count \\\midrule
Queensland & 4 \\
New South Wales & 4\\
Victoria & 4 \\
Northern Territory & 3 \\ 
Western Australia & 2 \\
\bottomrule
\end{tabular}
\caption{Location of interviewees.}
\label{tab:state_territory}
\end{table}

\end{document}